\documentclass{article}

\usepackage{arxiv}

\usepackage{hyperref}       
\usepackage{url}            
\usepackage{amsmath}       
\usepackage{amsfonts}       
\usepackage{graphicx}
\usepackage{natbib}
\usepackage{doi}
\usepackage{blindtext}
\usepackage[overload]{textcase}
\usepackage{algorithm2e}
\usepackage{xcolor}

\title{BaMANI: Bayesian Multi-Algorithm causal Network Inference}

\date{}

\author{\href{https://orcid.org/0000-0002-5410-3094}{\includegraphics[scale=0.06]{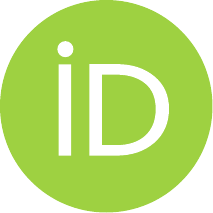}\hspace{1mm}Habibolla Latifizadeh}\thanks{Current affiliation: Department of Biostatistics and Bioinformatics, Duke University, Durham, NC.} \\
	School of Mathematical and Data Science\\
	West Virginia University\\
	Morgantown, WV 26501 \\
	\texttt{habibolla.latifizadeh@duke.edu} \\
	\And
	\href{https://orcid.org/0000-0002-5349-3561}{\includegraphics[scale=0.06]{orcid.pdf}\hspace{1mm}Anika C.~Pirkey} \\
	Department of Chemical and Biomedical Engineering\\
	West Virginia University\\
	Morgantown, WV 26501 \\
	\texttt{anika.pirkey@mail.wvu.edu} \\
	\And
	\href{https://orcid.org/0000-0001-7416-2799}{\includegraphics[scale=0.06]{orcid.pdf}\hspace{1mm}Alanna Gould} \\
	Department of Biochemistry\\
	The Medical College of Wisconsin\\
	Milwaukee, WI 53226 \\
	\texttt{algould@mcw.edu} \\
	\And
	\href{https://orcid.org/0000-0003-3299-4938}{\includegraphics[scale=0.06]{orcid.pdf}\hspace{1mm}David J.~Klinke II}\thanks{To whom correspondence should be addressed} \\
	Department of Chemical and Biomedical Engineering\\
	West Virginia University\\
	Morgantown, WV 26501 \\
	\texttt{david.klinke@mail.wvu.edu} \\
}

\renewcommand{\shorttitle}{BaMANI}

\hypersetup{
pdftitle={BaMANI: Bayesian Multi-Algorithm causal Network Inference},
pdfsubject={q-bio.NC, q-bio.QM},
pdfauthor={Habibolla Latifizadeh, Anika C. Pirkey, Alanna Gould, David J.~Klinke II},
pdfkeywords={Bayesian Network, Causal Inference, Blacklist, Whitelist, Structural Learning Algorithms},
}

\begin{document}
\maketitle

\begin{abstract}
Improved computational power has enabled different disciplines to predict causal relationships among modeled variables using Bayesian network inference. While many alternative algorithms have been proposed to improve the efficiency and reliability of network prediction, the predicted causal networks reflect the generative process but also bear an opaque imprint of the specific computational algorithm used. Following a ``wisdom of the crowds" strategy, we developed an ensemble learning approach to marginalize the impact of a single algorithm on Bayesian causal network inference. To introduce the approach, we first present the theoretical foundation of this framework. Next, we present a comprehensive implementation of the framework in terms of a new software tool called BaMANI (Bayesian Multi-Algorithm causal Network Inference). Finally, we describe a BaMANI use-case from biology, particularly within human breast cancer studies.
\end{abstract}

\keywords{Bayesian Network \and Causal Inference \and Blacklist \and Whitelist \and Structural Learning Algorithms}

\section{Introduction}
Understanding causal relationships among variables is of paramount importance in the fields of medicine, epidemiology, economics, social sciences, and beyond, where discerning causation from observations can have profound implications for policymaking, intervention strategies, and scientific discovery. The field of causal inference is dedicated to unearthing these relationships, providing methodologies to disentangle the intricate network of causality underlying observational data \citep{Pearl2009Causality, Spirtes2010Introduction}.

The seminal work by \cite{Pearl2009Causality} laid the groundwork for causal inference, introducing the do-calculus and structural causal models that have since been foundational in the field. \cite{Spirtes2010Introduction} further expand these concepts, providing robust methods for causal discovery from observational data.  Approaches such as Granger causality elucidate temporal causal relationships \citep{granger1969causality}. For data sets unstructured in time, Bayesian causal networks, leveraging the Markov Condition, facilitate a granular understanding of causal interactions \citep{Pearl2009Causality, glymour1999computation}. However, causal inference is challenged not only by conceptual and methodological complexities but also by computational demands. The inferential processes involved often require considerable computational resources, scaling with the size and complexity of the data sets \citep{Heckerman1995Learning, Chickering2002Optimal}.

In parallel with data sets increasing in size, advances in machine learning and statistics have further pushed the boundaries of computational efficiency in structure learning and causal discovery. In providing a historical perspective, Heckerman, Chickering, and Geiger \citep{Heckerman1995Learning, Chickering2002Optimal} discuss various algorithms designed to mitigate these computational demands, emphasizing the need for efficient and scalable solutions. More recently, an array of different algorithms have been developed. For instance, the Grow-Shrink Markov Blanket (GS) and  Incremental Association Markov Blanket (IAMB) algorithms are both geared towards determining the Markov blanket of a target variable \citep{Margaritis2003_thesis, Tsamardinos2003}. The Incremental Association with False Discovery Rate Control (IAMB.FDR), which extends the IAMB approach, integrates false discovery rate control to reduce erroneous causal links \citep{Pena2008, Gasse2014}. The Practical Constraint (PC.STABLE) algorithm \citep{ColomboMaathuis2014} has improved stability in large datasets, addressing one of the key limitations of the original PC algorithm.  Hill Climbing (HC) and Tabu Search (Tabu) \citep{RussellNorvig2009} algorithms adopt heuristic strategies and have offered innovative solutions for model refinement and exploration. The MMPC is an essential approach that identifies a set of potential parent and child nodes for each variable based on their mutual information, ensuring the initial structure is as accurate as possible \citep{Tsamardinos2003}. The Max-Min Hill-Climbing (MMHC) \citep{Tsamardinos2006} algorithm integrates constraint-based and score-based methods, starting with identifying candidate parents and children using the MMPC algorithm, followed by a hill-climbing search restricted to these candidates. Rounding off these methods, the Restricted Maximization (RSMAX2) optimizes the model's score while managing complexity \citep{Friedman1999}, and the Stable and Interpretable HITON Parents and Children (SI.HITON.PC) approach enhances the HITON algorithm for more stable and interpretable results, focusing on finding parent and child nodes that consistently appear across various data subsamples \citep{Aliferis2010}.

While available algorithms provide a diverse set of strategies, relying on a single algorithm for network inference can inadvertently reflect not only the underlying causal structure but also the biases inherent in an algorithm, which leads to an incomplete or skewed representation of the causal structure encoded in the data. To address this, we advocate for a ``wisdom of the crowds" framework, hypothesizing that an ensemble method could mitigate the effects of biases inherent in individual algorithms on the inferred causal network. Here, we introduce the BaMANI (Bayesian Multi-Algorithm Network Inference) framework, which leverages the strengths of multiple algorithms while attenuating their biases to yield a more comprehensive and accurate representation of the underlying causal structure, particularly in complex and high-dimensional data sets.

In the following sections, we outline the theoretical foundations of this ensemble approach before delving into a detailed implementation within BaMANI. We then showcase this framework through a case study using human breast cancer data. Our ensemble method not only integrates the strengths of individual algorithms, it also fosters a more robust and comprehensive model for causal network inference. This integration is crucial for tackling the inherent complexities of causal inference across diverse scientific fields, marking a step forward from existing methodologies that either rely on single network learning algorithms, such as bnlearn package \citep{Scutari2010}, or averaging techniques such as the ``wisdom of the crowds" approach  \citep{Marbach2012}.

\section*{Problem formulation}

The goal of causal inference is to uncover the underlying causal structure from observational data, represented as a Directed Acyclic Graph (DAG), \( G = (\mathcal{V}, \mathcal{E}) \). This graph encapsulates causal relationships among a set of variables \( \mathcal{V} = \{\mathcal{F}_1, \mathcal{F}_2, \cdots \mathcal{F}_{N_{f}}\} \), each corresponding to a distinct measurable feature/attribute within the observed data set \( \mathcal{D} \), comprising \( n \) instances across these variables. The task is to construct a DAG where the edges \( \mathcal{E} \subseteq \mathcal{V} \times \mathcal{V} \) symbolize direct causal influences, subject to critical constraints and assumptions intrinsic to causal discovery.

The DAG must inherently satisfy the principle of acyclicity, precluding any possibility of feedback loops where a variable might indirectly influence itself. This acyclic nature is fundamental to the concept of causality, as it adheres to the temporal precedence of cause over effect. Furthermore, the graph should conform to the statistical criterion of conditional independence. This criterion stipulates that any pair of non-adjacent variables in \( \mathcal{G} \) must be statistically independent, conditional on a set of other variables, if and only if they are d-separated within the graph. D-separated refers to determining if two sets of variables are conditionally independent by controlling for a third set of variables in a causal graph. This statistical dependence or independence is deciphered through rigorous probabilistic tests and is pivotal in determining the presence or absence of causal links.

The methods employed in constructing this causal graph are multifaceted. Score-based approaches involve an optimization task, where a scoring function, \( \text{score}(\mathcal{G}, \mathcal{D}) \), evaluates the fit of a DAG with respect to the data, seeking a graph structure that maximizes this score. In contrast, constraint-based methods rely on discerning conditional independencies directly from the data, using statistical tests to iteratively construct a graph that aligns with these independencies. Hybrid methods leverage the strengths of both score-based and constraint-based paradigms to attain a more robust and accurate causal model.

The resultant DAG, \( \mathcal{G} \), embodies the intricate network of causal relationships among the variables inferred from the data. This formulation not only delineates the mathematical and probabilistic underpinnings of causal inference but also underscores the complexity and nuances involved in extracting causality from empirical data. The endeavor extends beyond mere statistical correlation, delving into the realm of causation - a quest fundamental to understanding and modeling the dynamics of complex systems. As mentioned above, many structure learning algorithms aimed at causal learning have been developed. In this paper, we employ an array of such algorithms and build upon them via an ensemble approach with the goal of finding the most reliable DAG given the observational data.

\section*{BaMANI automated framework}
In this section, we will present the core idea of our ensemble causal network inference approach. This algorithm leverages the power of ensemble learning and minimizes the impact of a single algorithm on the network to achieve a more accurate estimation of the causal network. Below, we formulate the algorithm step-wise.

\subsection*{Algorithm Description}
The algorithmic representation of the BaMANI approach is presented in Algorithm \ref{alg:BNI}. As applied in practice, the R package code is published on GitHub \footnote{https://github.com/latifizadehhabib/BaMANI.Package}. This algorithm introduces a four-step process for inferring causal structure networks by leveraging domain knowledge, denoted as \( \mathcal{C} \), and observational data across features or nodes, denoted as \( \mathcal{D} \). The algorithm consists of the following four main steps, each optimizing a specific aspect of the knowledge integration and DAG learning process to provide an accurate and robust representation of the underlying causal dynamics. 

Given the data $\mathcal{D}= (\mathcal{D}_{\mathcal{F}_1}, \mathcal{D}_{\mathcal{F}_2}, \cdots , \mathcal{D}_{\mathcal{F}_{N_{f}}})$ where $\mathcal{D}_{\mathcal{F}_{i}}$ is features/nodes such that $\mathcal{D}_{\mathcal{F}_{i}}=\{ \mathcal{D}_{\mathcal{F}_{1i}}, \mathcal{D}_{\mathcal{F}_{2i}}, \cdots \mathcal{D}_{\mathcal{F}_{{n}_{obs}i}} \}$ represents the feature/node values of the $i$-th feature across all observations and set of directed arcs so-called blacklist $\mathcal{B}$ then a set of causal structure learning algorithms $ \mathcal{A}_1, \mathcal{A}_2, \cdots \mathcal{A}_n$ learn a set of corresponding DAG networks,  $ \mathcal{G}_1, \mathcal{G}_2, \cdots \mathcal{G}_n$ respectively. Here, $\mathcal{G}_i$ is defined as operator $\mathcal{A}_i(\mathcal{D}, \mathcal{B})$ and it has a set of directed arcs $\{ \mathcal{E}_{i1}, \mathcal{E}_{i2}, \cdots \mathcal{E}_{ik_i}\}$,  for $i=1, 2, \cdots, n$.

\RestyleAlgo{ruled}
\SetKwComment{Comment}{/* }{ */}

\begin{algorithm}[h]
\caption{BaMANI}\label{alg:BNI}
\KwData{Domain knowledge \( \mathcal{C} \), feature/node values \( \mathcal{D} \) for all observations/samples}
\KwResult{Inferred causal structure networks \( \mathcal{G} \) of the features}

\BlankLine

\textbf{Step 1: Create Blacklist}\\
\Indp
I: \( \mathcal{B} \leftarrow \{ (u, v) \mid (u, v) \not\cong \mathcal{C} \} \)\\
II: \( \Lambda \leftarrow \{  \mathcal{F}_{v} \mid \frac{1}{n} \sum_{i=1}^{n} \textbf{1}_{\{\mathcal{F}_{iv} = 0\}} > \frac{1}{2} \} \)\\
\Indm

\BlankLine

\textbf{Step 2: Generating an Ensemble of Potential arcs}\\
\Indp
Input: Ensemble algorithms $\{\mathcal{A}_1, \mathcal{A}_2, \cdots, \mathcal{A}_n\}$ with bootstrapped arc strength calculations\\
Initialize: $\mathcal{E} \leftarrow \emptyset$, $\mathcal{S} \leftarrow \emptyset$\\
\For{$i = 1$ \KwTo $n$}{
    $\mathcal{G}_i = \mathcal{A}_i(\mathcal{D}, \mathcal{B})$, where it has a list of arcs $\mathcal{E}_{i} = \{ \mathcal{E}_{i1}, \mathcal{E}_{i2}, \ldots, \mathcal{E}_{ik_i}\}$ each with strength $\mathcal{S}_{i} = \{ \mathcal{S}_{i1}, \mathcal{S}_{i2}, \ldots, \mathcal{S}_{ik_i}\}$\\
    $\mathcal{E} \leftarrow \mathcal{E} \cup \mathcal{E}_{i}$\\
    $\mathcal{S} \leftarrow \mathcal{S} \cup \mathcal{S}_{i}$\\
}
\Indm

\BlankLine

\textbf{Step 3: Filtering and Whitelisting}\\
\Indp
Input: Let N be the number of quantiles for the range of arc strength.\\
\For{$j = 1$ \KwTo $N-1$}{
    $Q_j = \min(\mathcal{S}) + j \frac{\max(\mathcal{S}) - \min(\mathcal{S})}{N}$ (\( j \)-th quantile threshold)\\
    \( \mathcal{W}_{temp} \leftarrow \{ \cup \mathcal{E}_{i{k_i}} \mid \mathcal{S}_{i{k_i}} < Q_j, \forall i \} \)\\
    \( \mathcal{G}^{'}_{j} \leftarrow \mathcal{A}(\mathcal{D}, \mathcal{B}, \mathcal{W}_{temp}) \)\\
    \For{$k = 1$ \KwTo $N_{f}$}{
        \( p_{kj} = \{ \cup (\mathcal{F}, \mathcal{F}_{k}) \mid (\mathcal{F}, \mathcal{F}_{k}) \in \mathcal{G}^{'}_{j}, \forall \mathcal{F} \in \mathcal{V} \} \)\\
        \( BIC_{kj} = |p_{kj}| \cdot \log_{10}(n_{obs}) + \sum_{i=1}^{n_{obs}} | \hat{\mathcal{F}}_{ik} - \mathcal{F}_{ik} | \)\\
    }
}
\For{$k = 1$ \KwTo $N_{f}$}{
    \( \mathcal{E}' \leftarrow \mathcal{E}' \cup \{ p_{kj'} \mid j' = arg \min_{1 \leq j \leq N-1} BIC_{kj} \} \)\\
}
\eIf{$\mathcal{E}'$ \text{ create cycle }}{
    $\mathcal{W} \leftarrow \mathcal{E}' \setminus (\text{inconsistent Arcs}(\mathcal{E}')$)\\
}{
    $\mathcal{W} \leftarrow \mathcal{E}'$ \tcp*{No cycles detected}
}
\Indm

\BlankLine

\textbf{Step 4: Final DAG Learning}\\
\Indp
\( \mathcal{G} \leftarrow \mathcal{A}(\mathcal{D}, \mathcal{B}, \mathcal{W}) \) \tcp*{Infer network structure}
\ForEach{\(v \in \mathcal{V}(\mathcal{G})\)}{
    \ForEach{\(u \in \text{Pa}(v) \text{ in } \mathcal{G}\)}{
        \( \theta_{uv} \leftarrow \Theta(u, v, \mathcal{D}) \) \tcp*{Estimate parameters for each arc}
    }
}
\Indm
\end{algorithm}

\vspace{3.5mm}

In the first step, BaMANI incorporates existing domain knowledge into the network structure. This is achieved by creating a blacklist, denoted as \( \mathcal{B} \), which contains pairs of nodes \( (u, v) \)  that are deemed to be inconsistent with the established domain knowledge \( \mathcal{C} \) (Constraints). By including these constraints, the algorithm ensures that the resulting network adheres to known causal relationships and logical constraints. Additionally, the user has the flexibility to identify specific features as ``leaf nodes" (\( \mathcal{F}_{v} \)) where the majority of the observations are zeroes. This feature helps in preventing spurious connections in the network and enhances the interpretability of the learning results.

Following blacklist creation, the algorithm proceeds to generate a list of potential arcs by employing an ensemble of structural learning algorithms (\( \{\mathcal{A}_1, \mathcal{A}_2, \ldots, \mathcal{A}_n\} \)). These algorithms, which are state-of-the-art in the causal Bayesian network and DAG learning literature, learn independently a set of DAGs denoted as (\( \mathcal{G}_1, \mathcal{G}_2, \ldots, \mathcal{G}_n \)) by imposing the blacklist \( \mathcal{B} \) into the learning process. To generate statistics associated with network learning, bootstrap resampling was used, that is resampling the dataset with replacement to generate a synthetic dataset of similar size as the original and infer the network structure using the synthetic dataset, and resulted in generating 10,000 network structures. For each algorithm, an averaged network structure was then calculated from this collection of network structures, where the threshold for including an arc into the average network was automatically determined by each algorithm and was nominally 0.5. 

Each algorithm $\mathcal{A}_i$ in the ensemble operates on the dataset $\mathcal{D}$ and the blacklist $\mathcal{B}$ with bootstrap resampling to average across the different algorithms, producing a potential causal structure network $\mathcal{G}_i$ with a specific set of directed arcs \( \mathcal{E}_i \) and their associated strengths \( \mathcal{S}_i \) representing the confidence of a causal relationship between the nodes \citep{scutari2013identifying}. The strength of an arc corresponds to the probability that the observed partial correlation between two nodes, linked by an arc, is explained by random chance rather than a significant causal linkage, given the rest of the network. Specifically, arc strength is the p-value calculated using the exact t-test for a Pearson's correlation coefficient. These arcs and their associated strengths are then aggregated across all the algorithms, resulting in a comprehensive set of potential arcs (\( \mathcal{E} \)) and their strengths (\( \mathcal{S} \)). 

The learning process involves both directed and undirected algorithms. Directed algorithms are key in deducing the network's structure and causality, forming DAGs by considering prior knowledge and arc strength. In parallel, undirected algorithms are used for local discovery, providing additional evidence for the existence of edges between nodes and enhancing the overall understanding of network causality.

In the advanced schema of BaMANI, network refinement is executed by balancing regression accuracy with model complexity, culminating in the formulation of whitelisted arcs. This phase is initiated by quantifying the evidential strength of each arc. Then, specific thresholds for arc strength are established by \( N \) distinct quantiles, facilitating the construction of temporary whitelists \( \mathcal{W}_{temp} \) corresponding to each threshold. These temporary whitelists are constructed by including the arcs from each algorithm that have strengths below the specific threshold. Subsequently, the blacklist \( \mathcal{B} \), along with the temporary whitelist \( \mathcal{W}_{temp} \) are used as inputs to the structure learning algorithm \( \mathcal{A} \) to obtain a refined causal structure network \( \mathcal{G'}_{j} \) for each threshold category. 

To quantify the balance between regression accuracy and model complexity, each feature \( \mathcal{F} \) within the reconstructed network is subjected to a Bayesian Information Criterion (BIC) evaluation, \( BIC_{kj} \). The BIC combines the absolute discrepancies between observed and predicted feature values, expressed as a L1 loss function, and an ad hoc penalty associated with the number of parameters, that is the number of parental arcs, to quantify the compatibility of local parental network associated with a feature. Calculating the BIC for each feature simplifies determining the local parental network and minimizes the impact of features having different magnitudes and dynamic ranges within the data set, as the minimum value may occur at different threshold categories for different features. Basically, selecting the minimum BIC score helps select the most appropriate parental arcs for a feature that corresponds to the threshold categories. 
 
\begin{figure}[!ht]
  \begin{center}
    \includegraphics[width=0.75\textwidth]{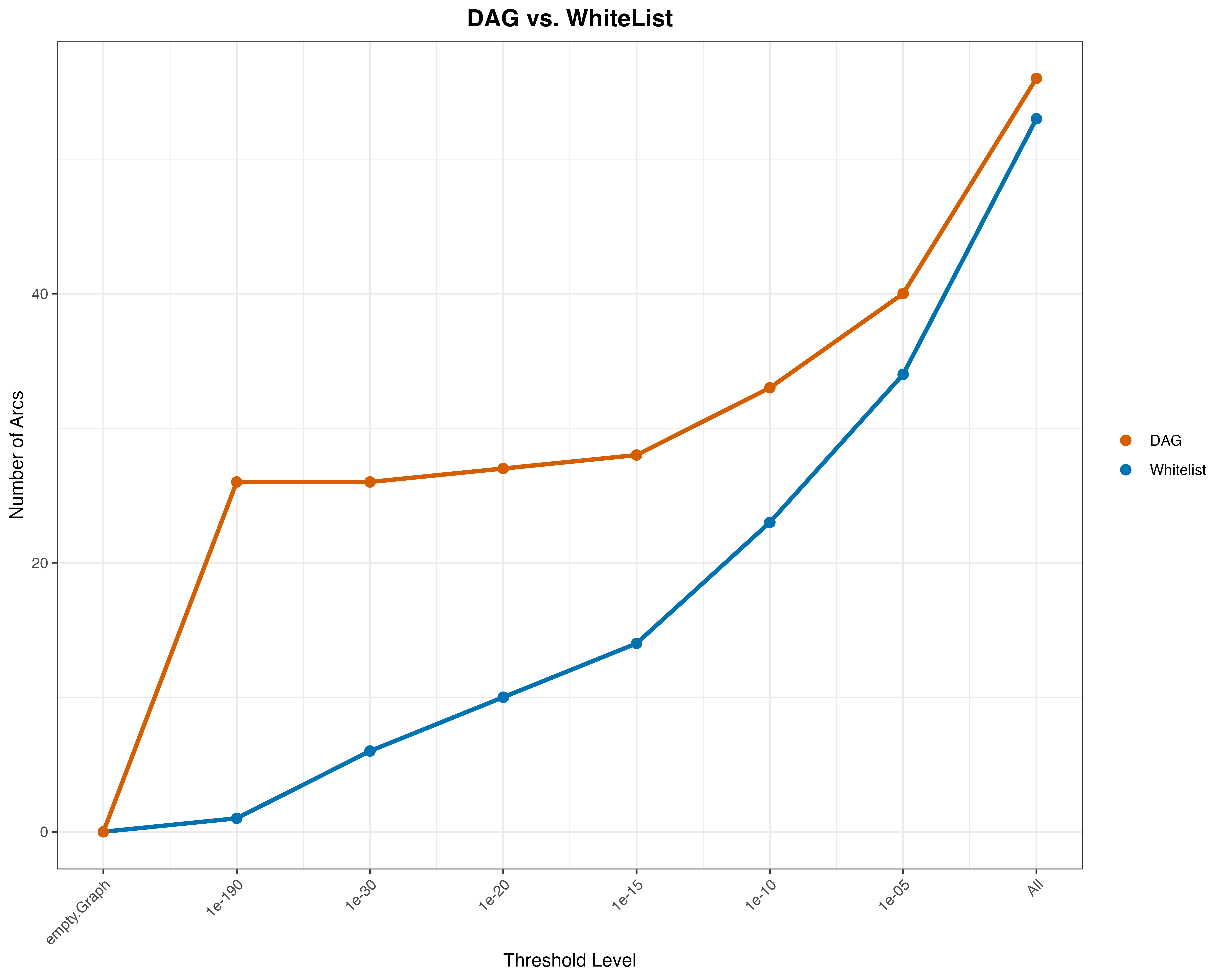}
  \end{center}
\caption{\textbf{Diagnostic plot.} Overall network connectivity (orange line) and number of arcs in the whitelist (blue line) as a function of edge strength threshold values (x-axis). \label{fig: Diagnostic_plot}}
\end{figure}

The next step involves aggregating arcs that correspond to the minimal BIC scores within their respective threshold categories. A critical component of this integration is the validation of acyclicity among these arcs, a measure imperative to maintaining the structural and functional integrity of the network, which can be checked by the user or automatically by BaMANI. The derived whitelist, \( \mathcal{W} \), serves as an exacting filter, selectively incorporating arcs substantiated by empirical evidence into the final network model. This prioritization of arcs for inclusion in the final network enhances interpretability and provides a more concise representation of the underlying causal dynamics.

The final learning stage of the BaMANI algorithm uses the collated data \( \mathcal{D} \), the blacklist \( \mathcal{B} \), and the whitelist \( \mathcal{W} \) from the last step to deduce the definitive network structure \( \mathcal{G} \). The resulting learned DAG was used to estimate parameters \( \theta_{uv} \) for a linear Gaussian model estimated by maximum likelihood and conditioned on the network structure that approximates the joint probability distribution associated with the data set. Values for the linear coefficients for each arc and the average node values were used to annotate the ensemble DAG. The sign of the linear coefficient was also used to annotate whether a particular parental arc promotes (black) or inhibits (red) the target child node. 

In summary, the algorithm combines domain expertise in providing blacklisted and whitelisted arcs with statistical and ensemble learning techniques. Additionally, we implement quantitative assessment through performance metrics such as arc strength and comparative network analysis with multiple individual algorithms to substantiate the robustness and accuracy of the resulting network. This combination ensures a precise and reliable reflection of the underlying causal dynamics inherent in the data. 

As network learning doesn't have to be a tabula rasa, BaMANI allows for the incorporation of domain expertise. For instance, users can designate nodes with a majority of zero values as leaf nodes. This means that when a feature named $X$ predominantly contains zero values, a list of arcs where $X$ is the cause (i.e., $X \to$ other column names) can be included in the blacklist. The DAG may still capture parents of $X$ but not consider whether $X$ causally influences other features. The rationale for specifying $X$ as a leaf node is that the majority presence of zero values suggests that the data set contains insufficient information about how $X$ varies with other features, that is the biological signals that the DAG intends to capture. Moreover, technical considerations may be driving whether X can be detected or not. If $X$ is not specified as a leaf node, incorporation of $X$ into the DAG may reflect technical consideration and not biological signals. In addition, designating the ``Cancer" node as root node represents basic biological understanding related to oncogenesis, that is genetic changes within epithelial cells within the mammary gland orchestrates the collective change in the cellular composition and function orientation within a tissue sample that a pathologist identifies as malignant.

This flexibility enables the algorithm to adhere to experts' understanding of causal relationships and provides a way to encode prior knowledge into the network structure. By using multiple structural learning algorithms, BaMANI reduces the bias inherent in any single algorithm and enhances the robustness of the inferred network structure. Performance metrics, including comparative network analysis with multiple individual algorithms and arc strength, are used in the ``Results and Validation" section to illustrate the robustness of BaMANI. The aggregated potential arcs and strengths enable the algorithm to identify robust causal relationships in the presence of noise and uncertainty. However, the reliability and interpretability of the inferred network strongly depend on the sample size and signal diversity of the data set used \citep{Kaiser2016}. When introducing new features to the network, it's crucial to ensure an adequate sample size to maintain confidence in the identified relationships or arcs within the network. Insufficient sample sizes can lead to unstable network structures and less interpretable results. Essentially, more sample data helps validate the robustness of the network's inferred connections, thereby enhancing the overall reliability and utility of the Bayesian network model in capturing the underlying data-driven relationships.

The interpretation of causal relationships by BaMANI can be substantially influenced by the specific data analyzed. Consequently, it is important that the data used for analysis be representative and free of biases that could result in unfair or discriminatory results. Although BaMANI offers valuable insights into potential causal connections, these findings are primarily causal hypotheses that require additional validation. It is essential to conduct experimental studies in order to validate these predictions and strengthen the credibility of the findings. Experts should also evaluate the outcomes and conclusions obtained from BaMANI to guarantee that they are consistent with established knowledge and ethical standards.

\section*{Results and validation}
\begin{figure}[!ht]
  \begin{center}
    \includegraphics[width=5in]{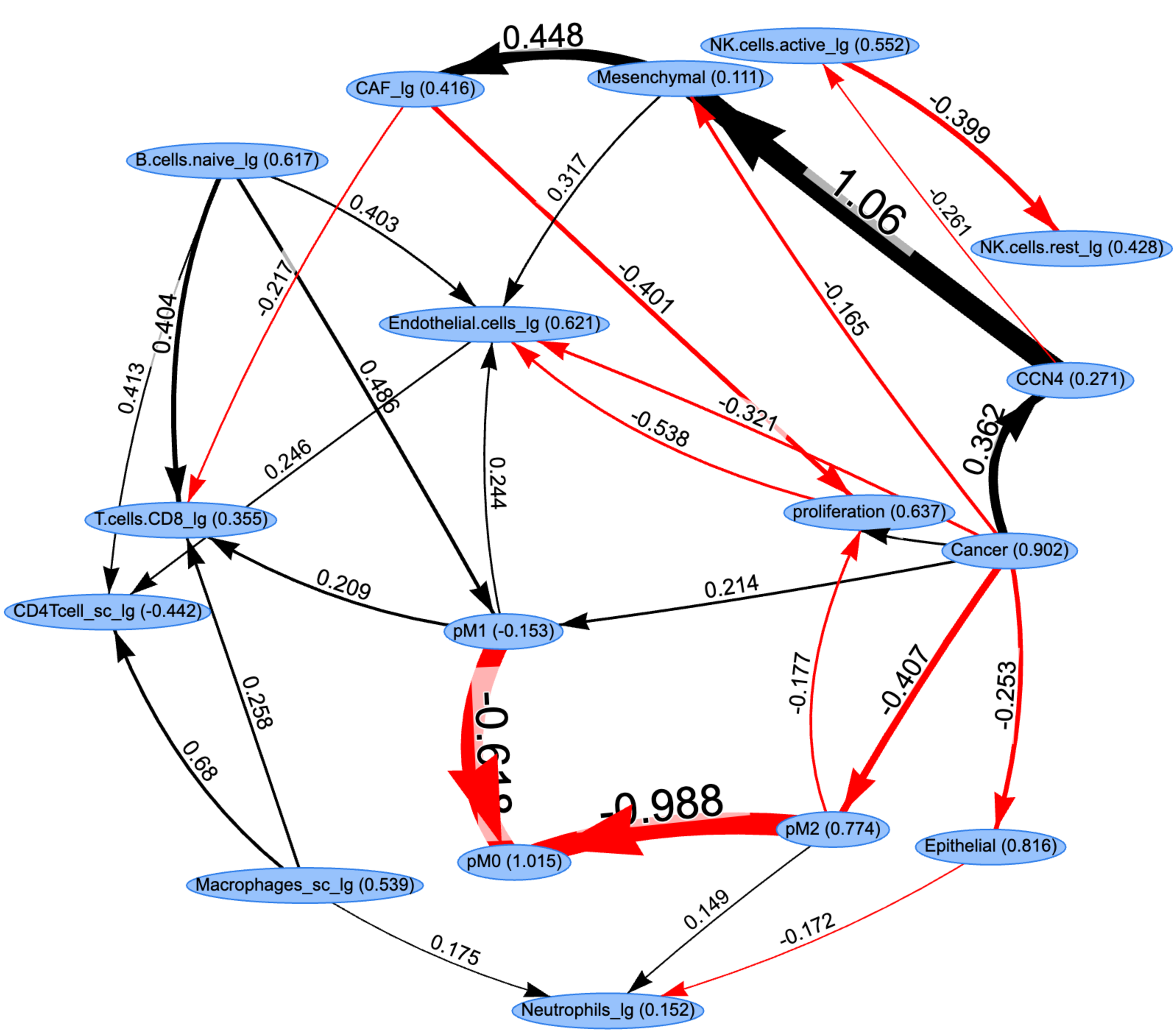}
  \end{center}
\caption{\textbf{BaMANI ensemble DAG for the Breast Cancer case study, pinpointing a nuanced understanding of the mechanistic interactions among the secreted protein CCN4 and cells within the tissue microenvironment.} \label{fig: Ensemble_DAG}}
\end{figure}

In this section, we present a comprehensive analysis of the causal relationships among cell types using various algorithms, comparing their performance with the BaMANI ensemble method. The algorithms analyzed include IAMB, IAMB.FDR, PC.STABLE, TABU, MMHC, GS, RSMAX2, and HC. For each algorithm, we examined the arcs representing causal relationships by comparing them to the arcs identified by the BaMANI ensemble DAG in Figure \ref{fig: Ensemble_DAG} which pinpoints a nuanced understanding of the mechanistic interactions among the secreted protein CCN4 and cells within the tissue microenvironment.

\begin{figure}[!ht]
  \begin{center}
    \includegraphics[width=6in]{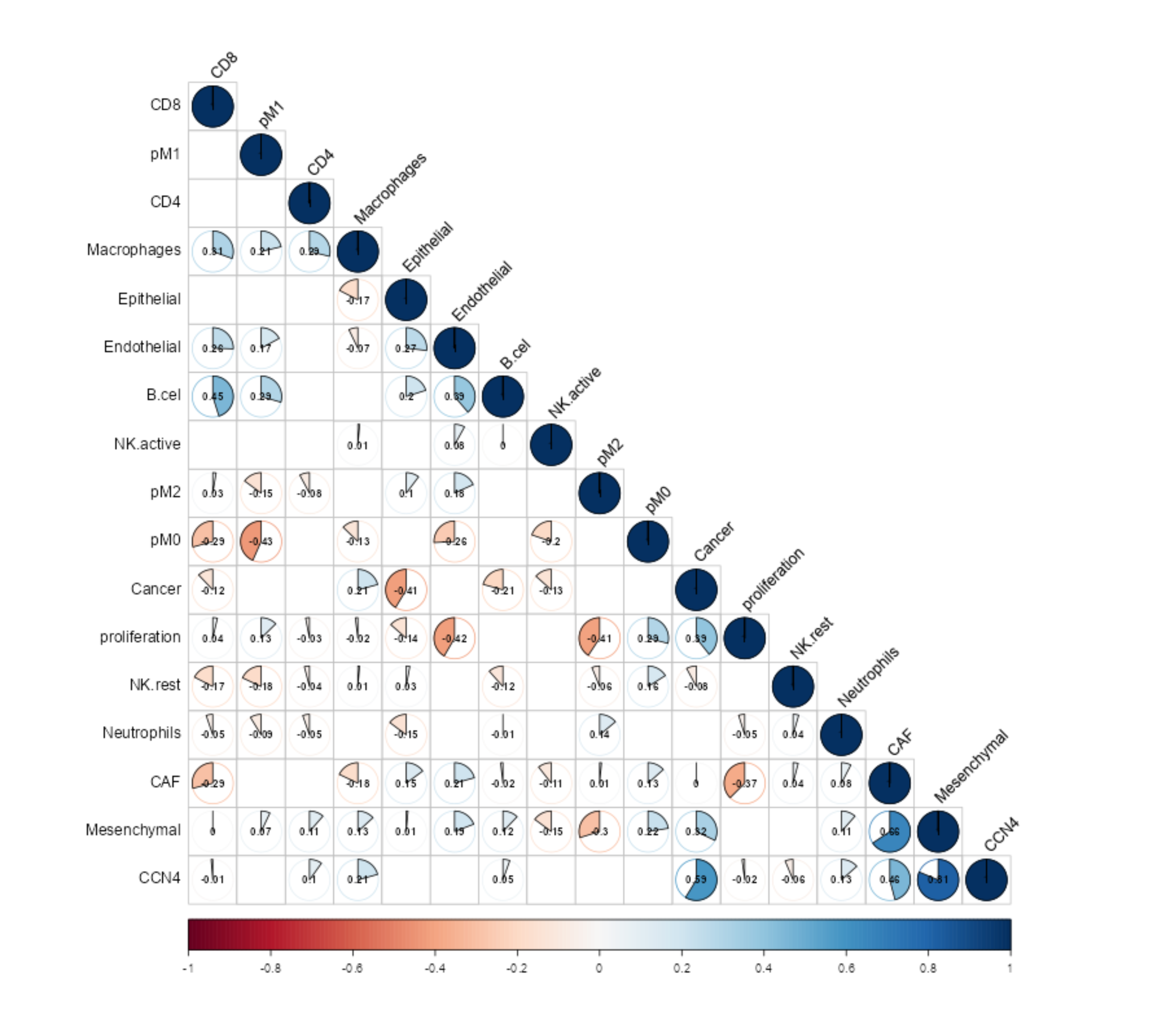}
  \end{center}
\caption{\textbf{Correlation matrix highlighting the relationships between CCN4 and Mesenchymal cell state, among other significant correlations in the breast cancer data.} \label{fig:significant_Correlation}}
\end{figure}

The application of BaMANI aimed to identify and validate causal relationships among these features, thus offering insights into the complex biological interactions driving breast cancer development. In particular, the correlation between CCN4 and Mesenchymal cell state, as highlighted in the correlation matrix in Figure \ref{fig:significant_Correlation}, is further interrogated within the causal DAG network. For instance, the DAG refines our understanding of the interactions involving CAFs and proliferation, M2 and proliferation, M1 and M0 with negative correlation by a direct arc, relation between B cells and CD8 T cells, and interaction of B cells with epithelial and endothelial cells with positive correlation by direct arc. In addition, the directed edges in the DAG not only corroborate this correlative association but also suggest a causal influence exerted by CCN4 on promoting a Mesenchymal cell state, which has been experimentally validated \citep{Deng2019}. Interestingly, the DAG includes an incoherent feedforward loop linking malignant transformation with CD8 T cells, a key cytotoxic effector cell that can eliminate malignant cells. The two sides of this incoherent feedforward loop include the Cancer $\to$ pM1 $\to$ T.cells.CD8 path that promotes CD8 T cells. The path Cancer $\to$ CCN4 $\to$ Mesenchymal $\to$ CAF $\to$ T.cells.CD8 is predicted to inhibit CD8 T cells.  

The predicted DAG transforms the observed correlations into actionable, causal insights that can be used to inform the design of experiments to further validate the existence of these relationships in biological systems. Moreover, controlling for CCN4 demonstrates that the direct effect of ``Cancer" on ``Mesenchymal" could be negative, as evidenced by the coefficient value of -0.165 in the Ensemble model, despite the positive direct correlation between the two variables. This is explained by Simpson's Paradox, which posits that the direction of an association can be reversed by the inclusion of additional variables. The relationship between ``Cancer" and ``Mesenchymal" is significantly influenced by the presence of CCN4 in the model, which underscores the limitation of correlation analysis in understanding complex biological interactions.

While there is no direct arc between Cancer-Associated Fibroblasts (CAFs) and Endothelial cells in the DAG network (Fig. \ref{fig: Ensemble_DAG}), the network suggests indirect causal pathways as indicated by significant correlation in the figure \ref{fig:significant_Correlation}. For instance, CAFs may influence Endothelial cells through intermediary nodes such as by inducing the proliferation of cells. One might postulate that an increased number of cells within a tissue creates a perfusion deficit and encourages the formation of new blood vessels via unmodeled mechanisms. By elucidating these indirect pathways, the BaMANI DAG network may refine our understanding of the interplay between CAFs and Endothelial cells beyond what the correlation shows. The directed edges in the DAG provide a mechanistic basis for these interactions, suggesting multi-step causal influences crucial for understanding the regulatory mechanisms within the tumor microenvironment. This mechanistic insight is pivotal for identifying potential therapeutic targets, as it highlights the importance of considering indirect effects and network-wide interactions when designing interventions.

\begin{figure}[!ht]
  \begin{center}
    \includegraphics[width=0.7\linewidth]{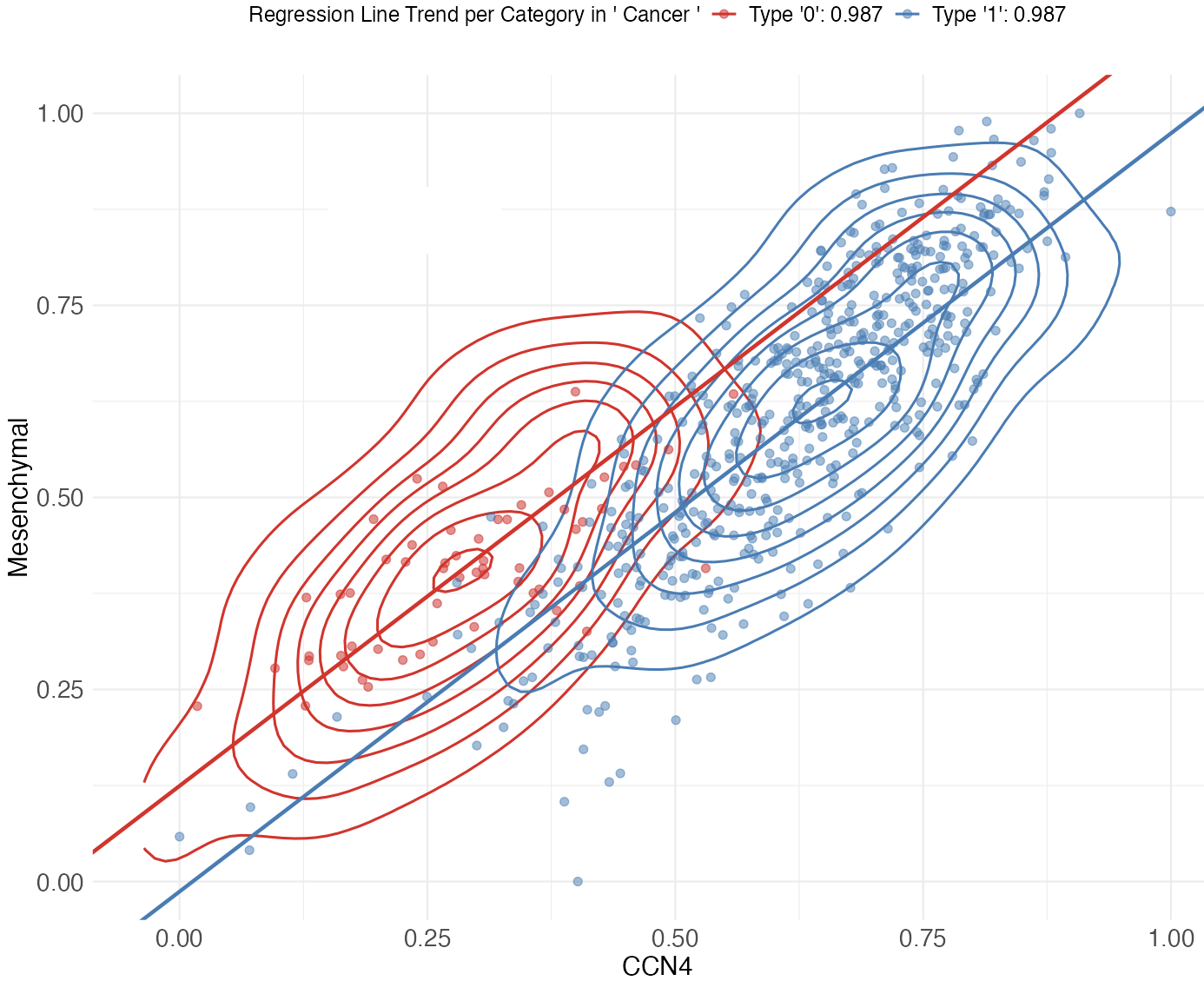}
  \end{center}
\caption{\textbf{Conditional probability query of the DAG.} Contour plots are colored red for cancer and blue for normal tissues, with corresponding dots representing experimental samples from these states. \label{fig: Contour_Plot}}
\end{figure}

\subsection*{Comparison}

To investigate the impact of CCN4 expression levels in normal and cancer tissues, we generated samples by querying the conditional probability distribution from the BaMANI ensemble DAG network. The contour plots in Figure \ref{fig: Contour_Plot} are colored red for cancer and blue for normal tissues, with corresponding dots representing experimental samples from these states. The posterior distributions reflect the experimental data points, indicating an increase in CCN4 expression from normal to cancer states. In comparing normal to cancer tissues, mesenchymal cells exhibited similar trends, with increased CCN4 expression showing a pronounced positive impact on their prevalence.

\begin{figure}[!ht]
  \begin{center}
    \includegraphics[width=1.05\textwidth]{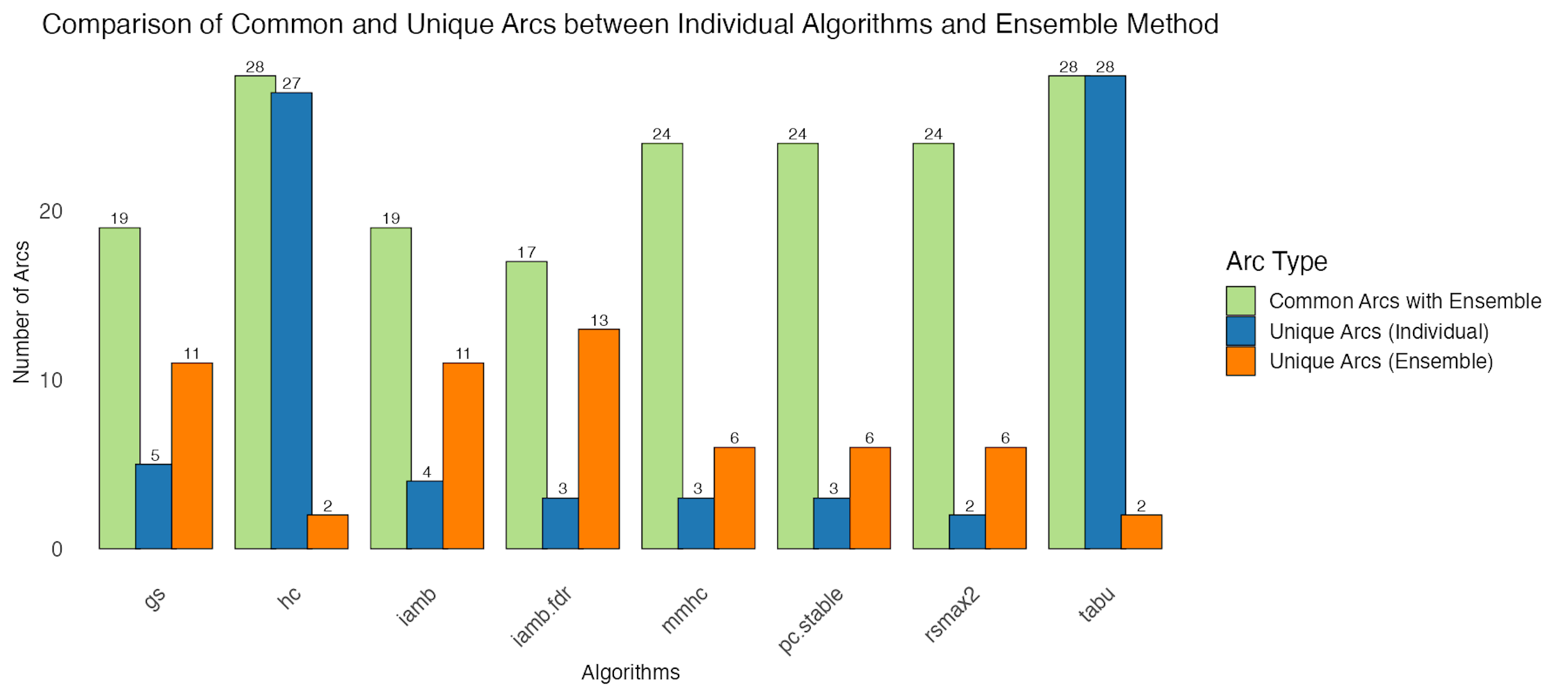}
  \end{center}
\caption{\textbf{Comparative Analysis of arcs from Ensemble DAG Network vs. Single Algorithm.} The plot shows the absolute counts of arcs for each algorithm, distinguishing between common arcs shared with the ensemble method (green), unique arcs identified by the individual algorithm (blue), and unique arcs identified by the ensemble (orange). \label{fig: Bar_plot}}
\end{figure}

\begin{figure}[!ht]
  \begin{center}
    \includegraphics[width=1.1\textwidth]{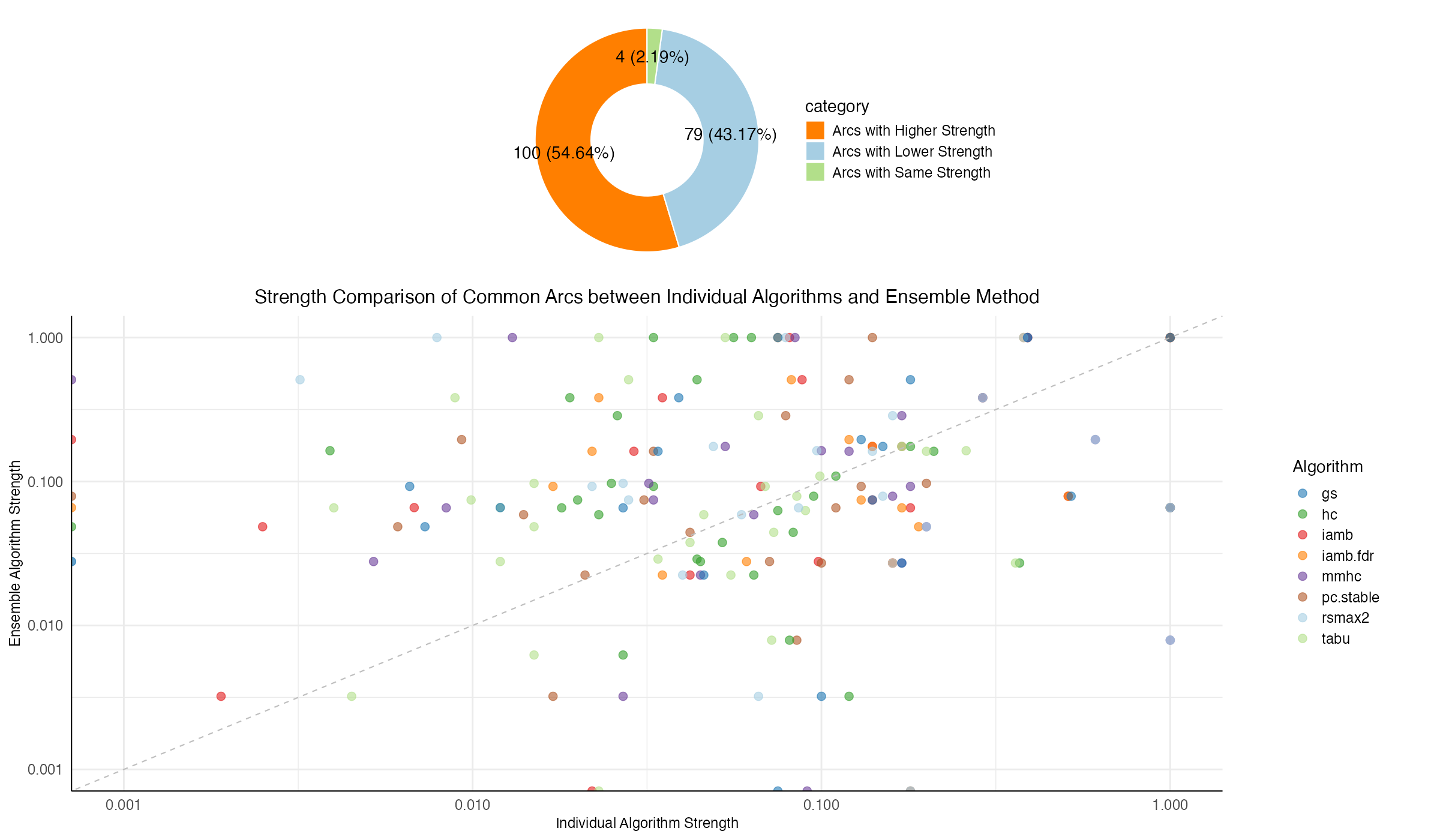}
  \end{center}
\caption{\textbf{Analyzing Common Arc Strength (log scale) in Ensemble vs. Single Algorithm.} The plot shows that points closer to the diagonal line (slope = 1) serving as a benchmark to indicate that the arc strength of individual algorithms matches closely with that of the ensemble. Points above the line indicate arcs that are stronger in the ensemble method, while points below indicate arcs stronger in the individual algorithm. \label{fig: Scatter_plot}}
\end{figure}

Additionally, the comprehensive comparative study of the results of the BaMANI ensemble algorithm have been visualized using a bar plot (Fig. \ref{fig: Bar_plot}) and scatter plot (Fig. \ref{fig: Scatter_plot}) to benchmark the performance of individual algorithms against the developed ensemble method. 

The bar plots in Fig. \ref{fig: Bar_plot} illustrate each algorithm's absolute counts of common and unique arcs. This visualization helps in understanding the number of arcs each individual algorithm uniquely contributes and the number of arcs each algorithm has in common with the ensemble method. From the bar plots, it is evident that some algorithms, like RSMAX2, MMHC, PC.STABLE, IAMB.FDR and IAMB contribute a significant number of common arcs to the ensemble method. This indicates that these algorithms often agree with the ensemble, suggesting their individual results are reliable and align well with the aggregated findings.

For a more detailed arc strength benchmark, the log-log scatter plot (Fig. \ref{fig: Scatter_plot}), compares the strengths of common arcs as determined by the individual algorithms against those determined by the BaMANI ensemble approach. The strength of an arc is measured by its arc strength value, which indicates the reliability of the relationship between nodes in the network. The scatter plots show that points closer to the diagonal line (slope = 1) serving as a reference to indicate that the arc strength of individual algorithms matches closely with that of the ensemble. Points above the line indicate arcs that are stronger in the ensemble method, while points below indicate arcs stronger in the individual algorithm. Additionally, the scatter plot was annotated with the percentage and number of arcs with higher, lower, or equal strengths compared to the ensemble method. The pie chart shows that 100 arcs (54.6\%) are identified with higher strength by the BaMANI ensemble algorithm, 79 arcs (43.2\%) are identified with higher strength by the individual algorithm, and 4 arcs (2.2\%) have equal strength in both the ensemble and the individual algorithms. This arc strength benchmark reveals that over half of the common arcs have higher strength in the ensemble algorithm, emphasizing the value of the ensemble algorithm in capturing causal relationships over randomly selecting any single algorithm used in isolation. 

\section*{Limitations and Future Work}
The analysis of the breast cancer data hinges on the assumption that tissue samples obtained from a cross section of patients diagnosed with breast cancer represent random samples drawn along a dynamic oncogenic process that is conserved across individuals. Insufficient sampling of this dynamic process - in terms of normal homeostasis; initiation; early, middle and late progression; and productive resolution or adverse outcomes - may skew the result. In addition, one of the limitations of inferring the network topology in the form of directed acyclic graphs is that indirect and direct causal relationships, such as modes of communication that involve feedback between cells, can be confounded. Collectively, predicted DAGs can motivate independent experiments that validate the predicted structure. One of the challenges with working with human data is that these independent controlled experiments may be difficult to conduct ethically. To overcome these ethical barriers, better understanding of the fidelity between pre-clinical experimental models and human oncogenesis is needed. Specifically, applying a multimodal workflow to tissue samples obtained along a defined disease trajectory would enable generating an analogous data set with corresponding ground truth measurements of cellular prevalence and function. Once the predicted pre-clinical DAGs were validated, obtaining a similar network topology with human oncogenesis would suggest similar biological mechanisms and help in selecting relevant pre-clinical models for testing novel therapies.

\section*{Conclusion}
In this paper, we introduced BaMANI, a novel ensemble approach for Bayesian causal network inference that leverages multiple algorithms to marginalize the impact of any single algorithm on the inferred network structure. By integrating the strengths of various algorithms and incorporating domain knowledge through blacklists and whitelists, BaMANI provides a more robust and accurate representation of causal relationships in complex systems.

The application of BaMANI to breast cancer data demonstrated its effectiveness in uncovering meaningful causal relationships among cell types and signaling molecules in the tumor microenvironment. The ensemble approach not only identified key causal pathways but also provided insights into the complex interactions that drive cancer progression, such as the role of CCN4 in promoting a mesenchymal cell state and the incoherent feedforward loop linking malignant transformation with CD8 T cells. Comparative analysis with individual algorithms showed that BaMANI consistently identified stronger causal relationships, with over half of the common arcs having higher strength in the ensemble algorithm compared to individual algorithms. This highlights the value of the ensemble approach in capturing robust causal relationships over relying on any single algorithm.

While BaMANI represents a significant advancement in causal network inference, it is important to recognize its limitations, particularly in the context of feedback loops and the need for experimental validation of predicted causal relationships. Future work should focus on extending the approach to handle cyclic relationships and developing methods for integrating experimental validation into the inference process. Overall, BaMANI provides a powerful tool for researchers seeking to uncover causal relationships in complex systems, offering a more comprehensive and reliable approach to causal network inference that can inform experimental design and advance our understanding of complex biological processes.

\end{document}